\documentclass{article}
\usepackage{times}
\usepackage{graphicx} % more modern
\usepackage{subfigure} 
\usepackage{natbib}
\usepackage{algorithm}
\usepackage{algorithmic}
\usepackage{hyperref}
\usepackage[fleqn]{amsmath}
\usepackage{amsfonts}

\usepackage[accepted]{icml2017}
\usepackage[none]{hyphenat}
\icmltitlerunning{A New Softmax Operator for Reinforcement Learning}

\newcommand{\sumizn}{\sum_{i=1}^n}
\DeclareMathOperator*{\E}{\mathbb E}
\DeclareMathOperator*{\argmax}{argmax}
\DeclareMathOperator*{\argmin}{argmin}

\icmltitlerunning{An Alternative Softmax Operator for Reinforcement Learning}
\begin{document} 

\twocolumn[
\icmltitle{An Alternative Softmax Operator for Reinforcement Learning}

%\icmlauthor{Kavosh Asadi}{kavosh@brown.edu}
%\icmladdress{Brown University}
%\icmlauthor{Michael L. Littman}{mlittman@brown.edu}
%\icmladdress{Brown University}

\icmlkeywords{boring formatting information, machine learning, ICML}

\begin{icmlauthorlist}
\icmlauthor{Kavosh Asadi}{br}
\icmlauthor{Michael L. Littman}{br}
\end{icmlauthorlist}

\icmlaffiliation{br}{Brown University, USA}

\icmlcorrespondingauthor{Kavosh Asadi}{kavosh@brown.edu}

\vskip 0.3in
]
\printAffiliationsAndNotice{}  
\begin{abstract} 
A softmax operator applied to a set of values acts somewhat like the maximization function and somewhat like an average. In sequential decision making, softmax is often used in settings where it is necessary to maximize utility but also to hedge against problems that arise from putting all of one's weight behind a single maximum utility decision. The Boltzmann softmax operator is the most commonly used softmax operator in this setting, but we show that this operator is prone to misbehavior. In this work, we study a differentiable softmax operator that, among other properties, is a non-expansion ensuring a convergent behavior in learning and planning. We introduce a variant of SARSA algorithm that, by utilizing the new operator, computes a Boltzmann policy with a state-dependent temperature parameter. We show that the algorithm is convergent and that it performs favorably in practice.
\end{abstract} 

\section{Introduction}
There is a fundamental tension in decision making between choosing the
action that has highest expected utility and avoiding ``starving'' the other actions. The issue arises in the context of the exploration--exploitation dilemma~\citep{Thrun92}, non-stationary decision problems~\citep{sutton90}, and when interpreting observed decisions~\citep{baker2007goal}.

In reinforcement learning, an approach to addressing the tension is the use of
\emph{softmax} operators for value-function optimization, and softmax policies for action selection. Examples include value-based methods such as SARSA \citep{rummery94b} or expected SARSA \citep{sutton98,van2009theoretical}, and policy-search methods such as REINFORCE \citep{Williams92}.

An ideal softmax operator is a parameterized set of operators that:
\begin{enumerate}
\item has parameter settings that allow it to
  approximate maximization arbitrarily accurately to perform reward-seeking behavior;
\label{p:max}
\item is a non-expansion for all parameter settings ensuring convergence to a unique fixed point;
\label{p:nonexpansion}
\item is differentiable to make it possible to improve via gradient-based optimization; and
\label{p:differentiable}
\item avoids the starvation of non-maximizing actions.
\label{p:nonstarving}
\end{enumerate}

\newcommand{\mean}{\mbox{\rm mean}}
\newcommand{\boltz}{\mbox{\rm boltz}}
\newcommand{\eps}{\mbox{\rm eps}}
\newcommand{\mm}{\mbox{\rm mm}}
\newcommand{\bX}{\textbf{X}}
\newcommand{\bY}{\textbf{Y}}

Let $\bX = x_1,\ldots,x_n$ be a vector of values. We define the following operators:
$$\max(\bX) = \max_{i \in \{1,\ldots,n\} } x_i\ ,$$
$$\mean(\bX) = \frac{1}{n}\; \sumizn x_i\ ,$$
$$\eps_\epsilon(\bX) = \epsilon \; \mean(\bX) + (1-\epsilon) \max(\bX)\ ,$$
$$\boltz_\beta(\bX) = \frac{\sumizn x_i\ e^{\beta x_i}}{\sumizn e^{\beta x_i}}\ .$$
The first operator, $\max(\bX)$,
is known to be a
non-expansion~\citep{littman96}. However, it is non-differentiable (Property~\ref{p:differentiable}), and ignores
non-maximizing selections (Property~\ref{p:nonstarving}).

% A greedy agent that performs pure
% maximization can converge to an inferior policy.

The next operator, $\mean(\bX)$, computes the average of its inputs. It is differentiable and, like any operator that takes
a fixed convex combination of its inputs, is a non-expansion. However, it does not allow for maximization (Property~\ref{p:max}).

The third operator $\eps_\epsilon(\bX)$, commonly referred to as epsilon
greedy~\citep{sutton98}, interpolates between $\max$ and $\mean$. The
operator is a non-expansion, because it is a convex combination of two non-expansion operators. But it is non-differentiable (Property~\ref{p:differentiable}).

The Boltzmann operator $\boltz_\beta(\bX)$ is
differentiable. It also approximates $\max$ as $\beta
\rightarrow \infty$, and $\mean$ as $\beta \rightarrow 0$.  However,
it is not a non-expansion (Property~\ref{p:nonexpansion}), and therefore, prone to misbehavior as will be shown in the next section.

In the following section, we provide a simple example illustrating why the non-expansion property is important, especially in the context of planning and on-policy learning. We then present a new softmax operator that is similar to the Boltzmann operator yet is a non-expansion. We prove several critical properties of this new operator, introduce a new softmax policy, and present empirical~results.

\section{Boltzmann Misbehaves}

\begin{figure}
\centering
\includegraphics[width=0.55\columnwidth]{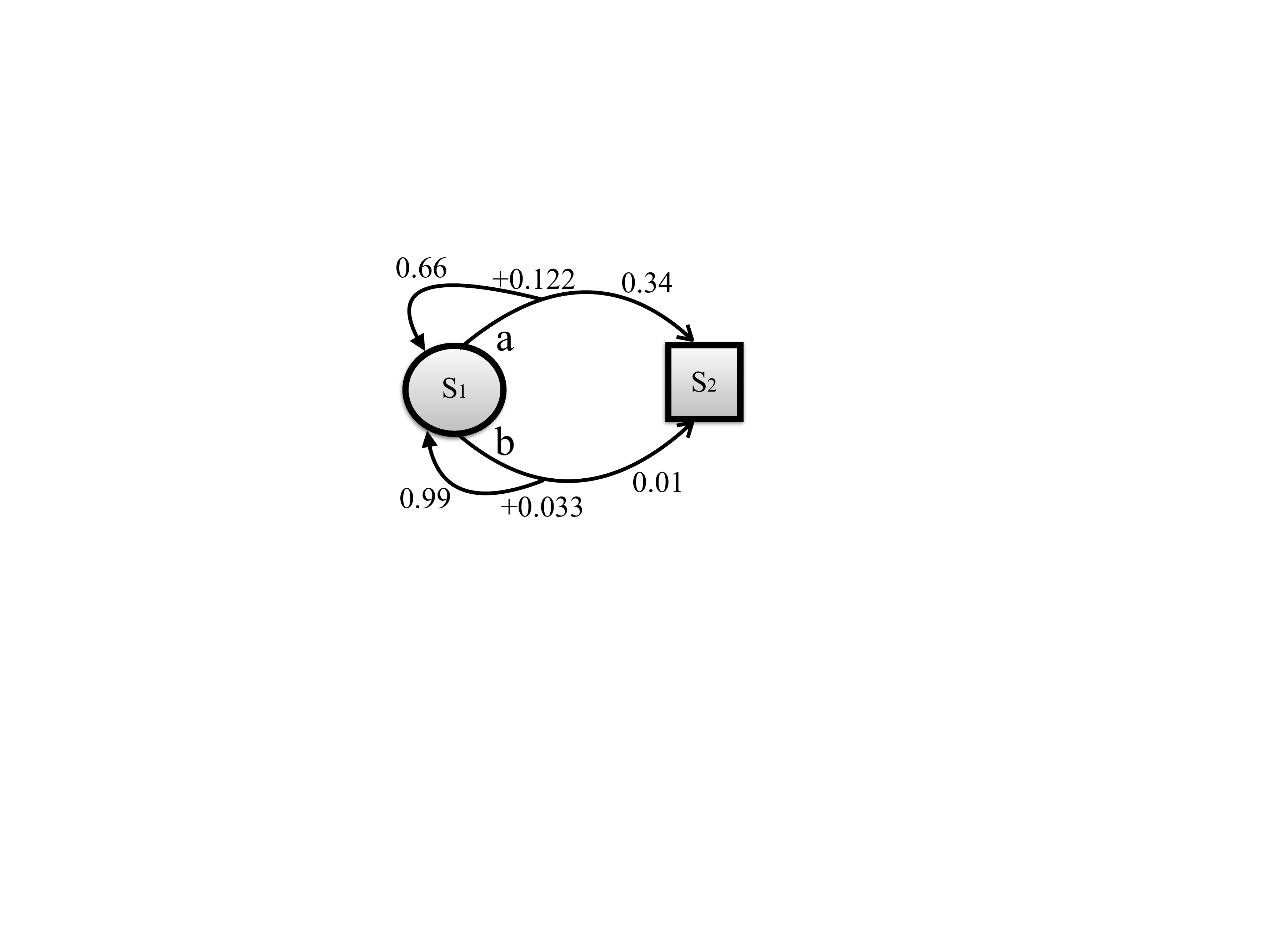}
\caption{A simple MDP with two states, two actions, and
  $\gamma~=~0.98\ $. The use of a Boltzmann softmax policy is not sound in this simple domain.}
\label{f:ambig}
\end{figure}

We first show that $\boltz_\beta$ can lead to problematic behavior. To this end, we ran SARSA with Boltzmann softmax policy (Algorithm \ref{alg:SARSABoltz}) on the MDP shown in Figure~\ref{f:ambig}. The edges are labeled with a transition probability (unsigned) and a reward number (signed). Also, state $s_2$ is a terminal state, so we only consider two action values, namely $\hat{Q}(s_1,a)$ and $\hat{Q}(s_2,b)$.
Recall that the Boltzmann softmax policy assigns the following probability to each action: 
\begin{equation*}
\pi(a|s)=\frac{e^{\beta \hat Q(s,a)}}{\sum_a e^{\beta \hat Q(s,a)}}\ .
\end{equation*}

\begin{algorithm}
\begin{algorithmic}
   \STATE {\bfseries Input:} initial $\hat Q(s,a)\ \forall s\in \mathcal{S}\ \forall a \in \mathcal{A}$, $\alpha$, and $\beta$
   \FOR{each episode}
   \STATE Initialize $s$
   \STATE $a \sim$ Boltzmann with parameter $\beta$
   \REPEAT
    \STATE Take action $a$, observe $r,s^\prime$
    \STATE $a^{'} \sim$ Boltzmann with parameter $\beta$
    \STATE $\hat{Q}(s,a) \leftarrow \hat{Q}(s,a) + \alpha\Big[r+\gamma \hat{Q}(s',a')-\hat{Q}(s,a)\Big]$
    \STATE $s \leftarrow s^{'}, a \leftarrow a^{'}$
   \UNTIL{$s$ is terminal}
  \ENDFOR

\end{algorithmic}
 \caption{SARSA with Boltzmann softmax policy}
 \label{alg:SARSABoltz}
\end{algorithm}

\begin{figure}
\centering
\includegraphics[width=0.6\columnwidth]{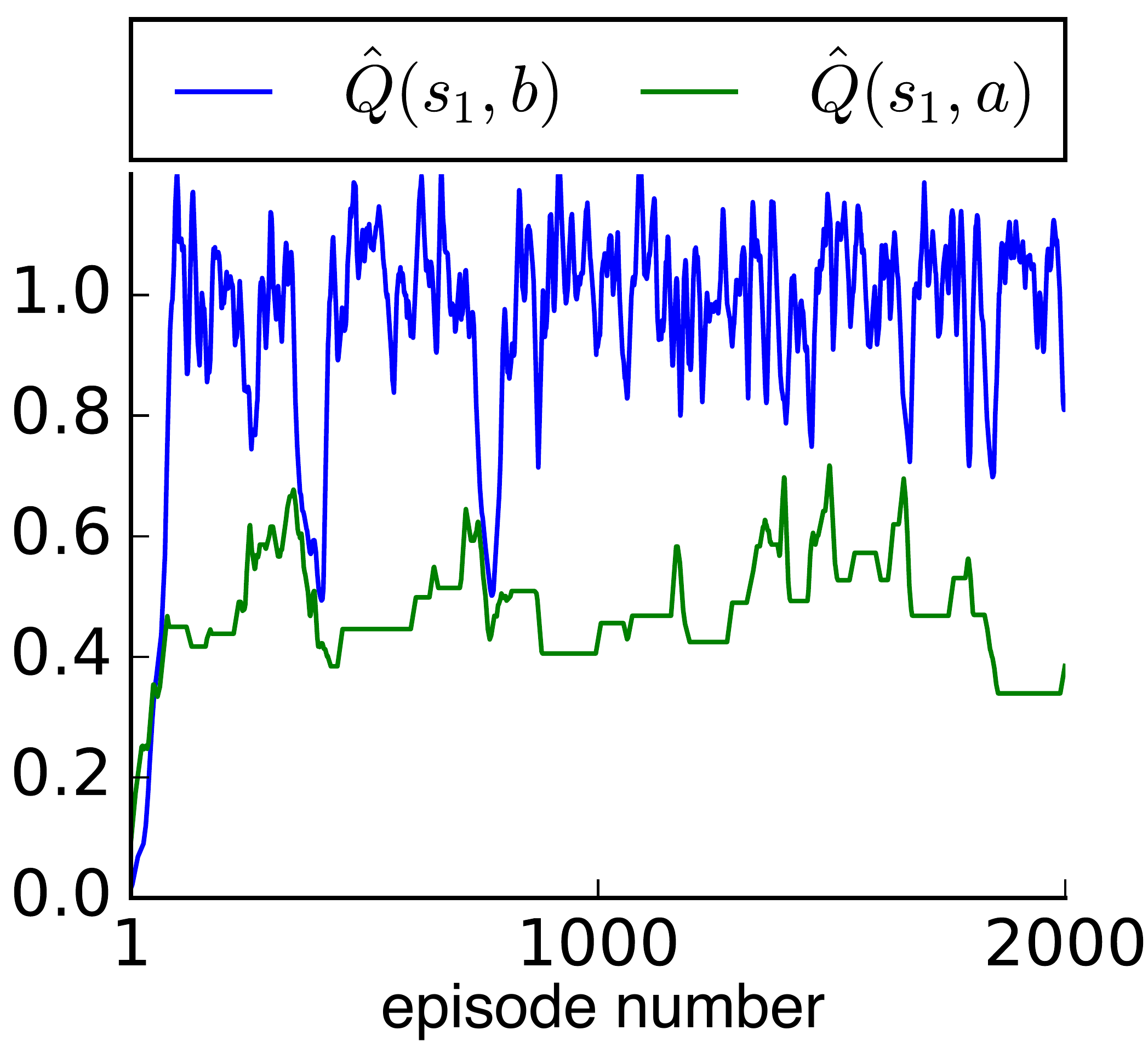}
\caption{Values estimated by SARSA with Boltzmann softmax. The algorithm never achieves stable values.}
\label{f:sarsaInnocentBoltz}
\end{figure}

In Figure~\ref{f:sarsaInnocentBoltz}, we plot state--action value estimates at the end of each episode of a single run (smoothed by averaging over ten consecutive points). We set $\alpha=.1$ and $\beta=16.55$. The value estimates are unstable.

SARSA is known to converge in the tabular setting using $\epsilon$-greedy exploration~\citep{littman96}, under decreasing exploration~\citep{singh00}, and to a region in the function-approximation setting~\citep{gordon2001reinforcement}. There are also variants of the SARSA update rule that converge more generally~\citep{perkins2002convergent,baird1999gradient,van2009theoretical}. However, this example is the first, to our knowledge, to show that SARSA fails to converge in the tabular setting with Boltzmann policy. The next section provides background for our analysis of the example.
\section{Background}

A Markov decision process~\citep{Puterman94}, or MDP,  is specified by the tuple $\langle \mathcal{S,A,R,P,\gamma}
\rangle$, where $\mathcal{S}$ is the set of states and
$\mathcal{A}$ is the set of actions. The functions $\mathcal{R:S\times A\rightarrow}\ \mathbb R$ and
$\mathcal{P:S\times\ A \times S\rightarrow}\ [0,1]$ denote the reward and transition dynamics of the
MDP.
% More precisely, the expected reward following an action
% $a \in \mathcal{A}$ in a state $s \in \mathcal{S}$ upon moving to a
% next state $s' \in \mathcal{S}$ is specified by:
% $$\mathcal{R}(s,a,s')= \E [R_{t+1}  |S_t=s,A_t=a,S_{t+1}=s']\ ,$$
% $ and the probability of this transition is defined by:
% $$\mathcal{P}(s,a,s')= \textrm{Pr}(S_{t+1}=s'\Big|S_t=s,A_t=a)\ .$$ 
Finally, $\gamma \in [0,1)$, the
discount rate, determines the relative importance of immediate reward
as opposed to the rewards received in the future.

A typical approach to finding a good policy is to estimate
how good it is to be in a particular state---the state value
function. The value of a particular state $s$ given a policy $\pi$ and initial action $a$ is
written $Q_{\pi}(s,a)$.
% formally defined to be:
% $$V_{\pi}(s)=\E_{\pi} \big[R_{t+1} +\gamma R_{t+2} + \gamma^2 R_{t+3}+ ...\ \big |S_t=s\big] .$$
% Similarly, we define the state--action value function as:
% $$Q_{\pi}(s,a)= \E_\pi \big[R_{t+1} +\gamma R_{t+2} + \gamma^2 R_{t+3}+\ ...\ \big |S_t=s,A_t=a\big] .$$
We define the optimal value of a state--action pair % $Q^{\star}(s,a)$ as:
$Q^{\star}(s,a)= \max_{\pi} Q_{\pi}(s,a) .$
It is possible to define $Q^{\star}(s,a)$ recursively and as a function of the optimal value of the other
state--action pairs:
\begin{equation*}
\label{eq:Bellman}
Q^{\star}(s,a)=\mathcal{R}(s,a)+\sum_{s'\in \mathcal{S}}\gamma\ \mathcal{P}(s,a,s') \max_{a'}Q^{\star}(s',a')\ .
\end{equation*}
Bellman equations, such as the above, are at the core of many reinforcement-learning algorithms such as Value Iteration \cite{VI}.
The algorithm computes the value of the best policy in an iterative fashion:
$$\hat Q(s,a) \leftarrow \mathcal{R}(s,a)+ \gamma\sum_{s'\in \mathcal{S}} \mathcal{P}(s,a,s') \max_{a'}\hat Q(s',a') .$$
Regardless of its initial value, $\hat Q$ will  converge to $Q^*$. 
% $\hat Q$ can then be used for decision making.

\citet{littman96} generalized this algorithm by replacing the $\max$ operator by any arbitrary operator  $\bigotimes$,
% $$q^{\star}(s,a)=\sum_{s'\in \mathcal{S}} \mathcal{R}(s,a,s')+\gamma\mathcal{P}(s,a,s') \bigotimes_{a'} q^{\star}(s',a').$$
resulting in the generalized value iteration (GVI) algorithm with the following update rule:
\begin{equation}\hat Q(s,a) \leftarrow \mathcal{R}(s,a)+\gamma \sum_{s'\in \mathcal{S}}\ \gamma\mathcal{P}(s,a,s') \bigotimes_{a'} \hat Q(s',a') .
\label{eqn:GVI_update} \end{equation}
\begin{algorithm}
\caption{GVI algorithm}
\begin{algorithmic}
   \STATE {\bfseries Input:} initial $\hat Q(s,a)\ \forall s\in \mathcal{S}\ \forall a \in \mathcal{A}$ and $\delta \in \mathcal{R^{+}}$
   \REPEAT
   \STATE $\textrm{diff} \leftarrow 0$
   \FOR{each $s\in \mathcal{S}$}
   \FOR{each $a\in \mathcal{A}$}
   \STATE $Q_{copy}\leftarrow \hat Q(s,a)$
   \STATE $\hat Q(s,a) \leftarrow \sum_{s'\in \mathcal{S}} \mathcal{R}(s,a,s')$\\$\hspace{1cm}+\ \gamma\mathcal{P}(s,a,s') \bigotimes \hat Q(s',.)$ 
   \STATE $\textrm{diff} \leftarrow \max\big\{\textrm{diff},|Q_{copy}-\hat Q(s,a)|\big\}$
   \ENDFOR
   \ENDFOR
   \UNTIL{$\textrm{diff}<\delta$}
\end{algorithmic}
\label{VI}
\end{algorithm}

Crucially, convergence of GVI to a unique fixed point follows if operator $\bigotimes$ is a non-expansion with respect to the infinity norm:
$$\Big|\bigotimes_a \hat Q(s,a) - \bigotimes_a \hat Q'(s,a)\Big|\leq \max_a \Big|\hat Q(s,a) - \hat Q'(s,a)\Big| ,$$
for any $\hat{Q}$, $\hat{Q}'$ and $s$.
\begin{figure}
\centering
\includegraphics[width=1\columnwidth]{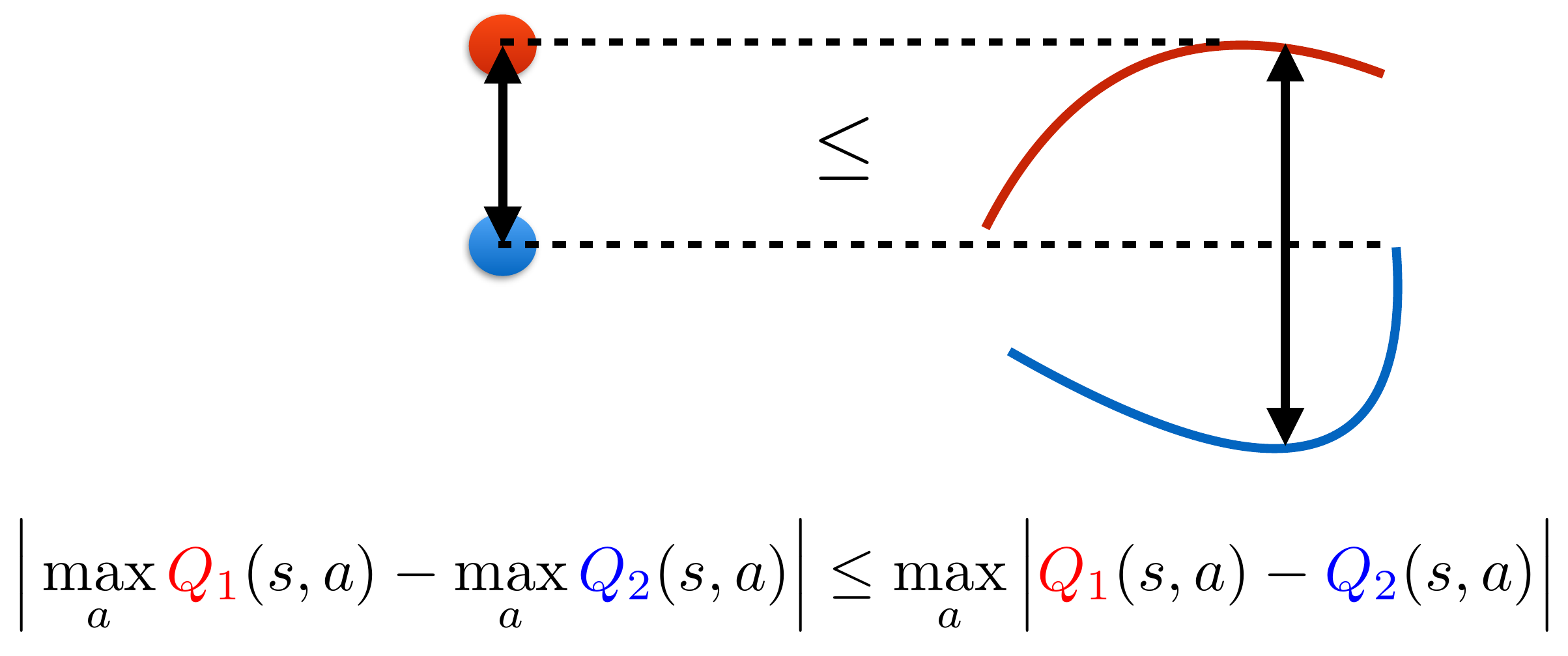}
\caption{$\max$ is a non-expansion under the infinity~norm. }
% While $\max$ is shown to be a non-expansion \citep{littman96c}, $\boltz_\beta$ is not a non-expansion.
\label{f:non_expansion_illustration}
\end{figure}
As mentioned earlier, the $\max$ operator is known to be a non-expansion, as illustrated in Figure~\ref{f:non_expansion_illustration}. $\mean$ and $\eps_\epsilon$ operators are also non-expansions. Therefore, each of these operators can play the role of $\bigotimes$ in GVI, resulting in convergence to the corresponding unique fixed point. However, the Boltzmann softmax operator, $\boltz_\beta$, is not a non-expansion~\citep{littman96c}. Note that we can relate GVI to SARSA by observing that SARSA's update is a stochastic implementation of GVI's update. Under a Boltzmann softmax policy $\pi$, the target of the (expected) SARSA update is the following:
\begin{eqnarray*}
\lefteqn{\E_\pi\big[r+\gamma \hat{Q}(s',a')\big|s,a\big]=}\\
&&\mathcal{R}(s,a)+\gamma\sum_{s' \in \mathcal S}\mathcal{P}(s,a,s')\underbrace{\sum_{a'\in\mathcal A}\pi(a'|s')\hat Q(s',a')}_{\boltz_{\beta}\big(\hat Q(s',\cdot)\big)}.
\end{eqnarray*}
This matches the GVI update (\ref{eqn:GVI_update}) when $\bigotimes = \boltz_\beta$.
\section{Boltzmann Has Multiple Fixed Points}
\begin{figure}
\centering
\includegraphics[width=1.0\columnwidth]{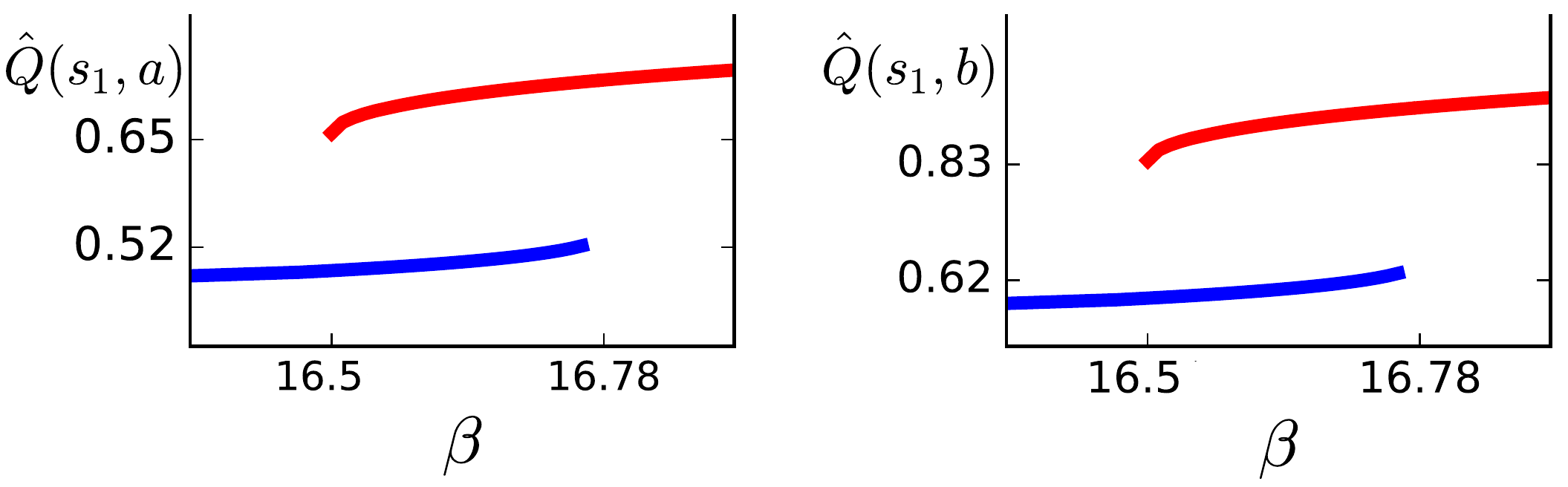}
\caption{Fixed points of GVI under $\boltz_\beta$ for varying $\beta$. Two distinct fixed points (red and blue) co-exist for a range of $\beta$.}
\label{f:fixedPoints}
\end{figure}

\begin{figure}
\centering
\includegraphics[width=.8\columnwidth]{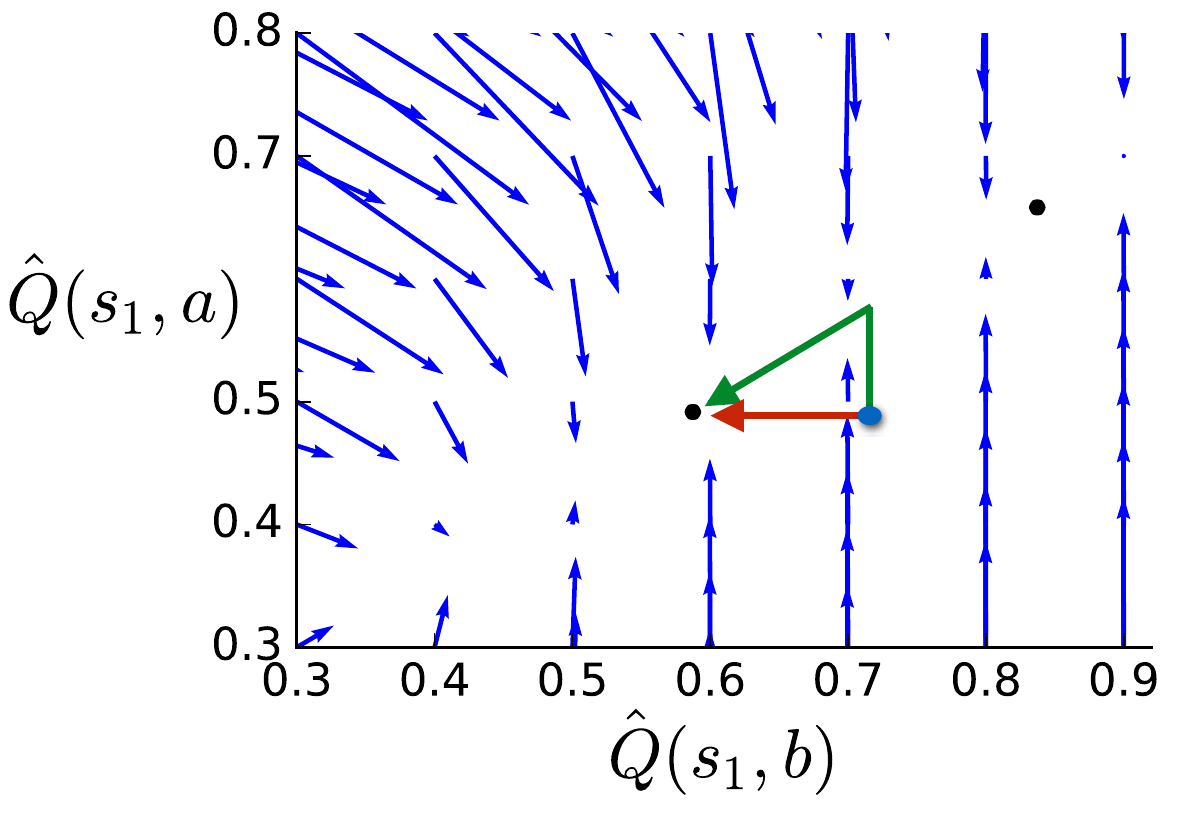}
\caption{A vector field showing GVI updates under $\boltz_{\beta=16.55}$. Fixed points are marked in black. For some points, such as the large blue point, updates can move the current estimates farther from the fixed points. Also, for points that lie in between the two fixed-points, progress is extremely slow.}
\label{f:vectorFieldBoltz}
\end{figure}

Although it has been known for a long time that the Boltzmann operator
is not a non-expansion~\citep{littman96c}, we are not aware of a
published example of an MDP for which two distinct fixed points
exist. The MDP presented in Figure \ref{f:ambig} is the first example where, as shown in Figure~\ref{f:fixedPoints}, GVI under $\boltz_\beta$ has two distinct fixed points. We also show, in Figure~\ref{f:vectorFieldBoltz}, a vector field visualizing GVI updates under $\boltz_{\beta=16.55}$. The updates can move the current estimates farther from the fixed points. The behavior of SARSA (Figure~\ref{f:sarsaInnocentBoltz}) results from the algorithm stochastically bouncing back and forth between the two fixed points. When the learning algorithm performs a sequence of noisy updates, it moves from a fixed point to the other. As we will show later, planning will also progress extremely slowly near the fixed points. The lack of the non-expansion property leads to multiple fixed points and ultimately a misbehavior in learning and planning.
% In this case, convergence point of GVI is determined by the initial value of the state-action values. This phenomena is due to $\boltz_\beta$ being an expansion, since GVI under a non-expansion operator is guaranteed to converge to a unique fixed point.

\section{Mellowmax and its Properties}

We advocate for an alternative softmax operator defined as follows:
$$\mm_\omega(\bX) =\frac{ \log (\frac{1}{n} \sumizn e^{\omega
    x_i})}{\omega}\ ,$$
which can be viewed as a particular instantiation of the quasi-arithmetic mean~\citep{beliakov16}. It can also be derived from information theoretical principles as a way of regularizing policies with a cost function defined by KL divergence~\cite{todorov2006linearly,rubin2012trading,fox2015taming}. Note that the operator has previously been utilized in other areas, such as power engineering \cite{logSumExpEE}.

We show that $\mm_\omega$,
which we refer to as \emph{mellowmax}, has the desired properties and that it compares quite favorably to $\boltz_\beta$ in practice. 
% We report empirical results on standard planning and reinforcement-learning domains, showing that the mellowmax operator can provide significant advantages.

\subsection{Mellowmax is a Non-Expansion}

We prove that $\mm_\omega$ is a non-expansion (Property \ref{p:nonexpansion}), and therefore, GVI and SARSA under $\mm_\omega$ are guaranteed to converge to a unique fixed point.

Let $\bX = x_1,\ldots,x_n$ and $\bY = y_1,\ldots,y_n$ be two vectors of values. Let $\Delta_i=x_i - y_i$ for $i \in \{1,\ldots,n\}$ be the difference of the $i$th components of the two vectors. Also, let $i^*$ be the index with the maximum component-wise difference, $i^*=\argmax_i \Delta_i$.  For simplicity, we assume that $i^*$ is unique and $\omega>0$. Also, without loss of generality, we assume that $x_{i^*} - y_{i^*} \geq 0$. It follows that: 
\begin{eqnarray*}
\lefteqn{\big|\mm_\omega(\bX)-\mm_\omega(\bY)\big|}\\
&=&  \big|\log (\frac{1}{n} \sumizn e^{\omega x_i})/\omega\ -
      \log (\frac{1}{n} \sumizn e^{\omega y_i})/\omega\ \big|\\
&=& \big|\log \frac{\frac{1}{n} \sumizn e^{\omega x_i}} {\frac{1}{n} \sumizn e^{\omega y_i}}/\omega\ \big|\\
&=& \big|\log \frac{\sumizn e^{\omega \big(y_i+\Delta_i\big)}} { \sumizn e^{\omega y_i}}/\omega\ \big|\\
&\leq&\big| \log \frac{\sumizn e^{\omega \big(y_i+\Delta_{i^*}\big)}} { \sumizn e^{\omega y_i}}/\omega\ \big| \\
\end{eqnarray*}
\begin{eqnarray*}
&=& \big|\log \frac{e^{\omega \Delta_{i^*}}\sumizn e^{\omega y_i}} { \sumizn e^{\omega y_i}}/\omega\ \big|\\
&=&\big|\log (e^{\omega \Delta_{i^*}})/\omega\big|
= \big|\Delta_{i^*}\big| 
% &=& \big|\max_{i} x_i-y_i\big|\\
= \max_i \big|x_i - y_i\big|\ ,
\end{eqnarray*}
 
allowing us to conclude that mellowmax is a non-expansion under the infinity norm.
\subsection{Maximization}

Mellowmax includes parameter settings that allow for 
maximization (Property~\ref{p:max}) as well as for minimization. In particular, as $\omega$ goes to infinity, $\mm_\omega$ acts like $\max$.

Let $m = \max(\bX)$ and let $W = |\{ x_i = m | i \in \{1, \ldots,
n\}\}|$.  Note that $W \geq 1$ is the number of maximum values (``winners'') in
$\bX$. Then:
\begin{eqnarray*}
\lim_{\omega\rightarrow \infty} \mm_\omega(\bX)
&=& \lim_{\omega\rightarrow \infty}  \frac{\log (\frac{1}{n} \sumizn e^{\omega x_i})}{\omega} \\
&=& \lim_{\omega\rightarrow \infty}  \frac{\log (\frac{1}{n} e^{\omega m} \sumizn e^{\omega (x_i-m)})}{\omega}\\
&=& \lim_{\omega\rightarrow \infty}  \frac{\log (\frac{1}{n} e^{\omega m} W)}{\omega} \\
&=& \lim_{\omega\rightarrow \infty}  \frac{\log (e^{\omega m}) -\log (n) + \log (W)}{\omega} \\
&=& m + \lim_{\omega\rightarrow \infty} \frac{-\log (n) + \log (W)}{\omega} \\
&=& m 
= \max(\bX)\  .
% $$ = \lim_{\omega\rightarrow \infty} \log ((\sum_{x\in W} e^{\omega x} + \sum_{x\in L} e^{\omega x})/|X|)/\omega$$
% $$ = \lim_{\omega\rightarrow \infty} \log (|W| e^{\omega m} + e^{\omega m} \sum_{x\in L} e^{\omega (x-m)})/|X|)/\omega.$$
% $$ = \lim_{\omega\rightarrow \infty} \log (|W| e^{\omega m} + e^{\omega m} \sum_{x\in L} e^{\omega (x-m)})/|X|)/\omega.$$
\end{eqnarray*}
That is, the operator acts more and more like pure maximization as the value of $\omega$ is increased. Conversely, 
as $\omega$ goes to $-\infty$, the operator approaches the minimum.
% \begin{eqnarray*}
% \lim_{\omega\rightarrow -\infty} \mm_\omega(\bX)
% &=& \lim_{\omega\rightarrow \infty} \mm_{-\omega}(\bX)\\
% &=& \lim_{\omega\rightarrow \infty} - \mm_{\omega}(-\bX)\\
% &=& -\max(-\bX)\\
% &=& \min(\bX)\ .
% \end{eqnarray*}

\subsection{Derivatives}
We can take the derivative of mellowmax with respect to each one of the arguments $x_i$ and for any non-zero $\omega$: 
\begin{equation*}
\frac{\partial \mm_\omega(\bX)}{\partial x_i}=\frac{e^{\omega x_i}}{\sumizn
  e^{\omega x_i}} \ge 0\ .
\label{e:nondecreasing}
\end{equation*}
 
Note that the operator is non-decreasing in each component of $\bX$.

Moreover, we can take the derivative of mellowmax with respect to $\omega$. We define $n_\omega(\bX)=\log (\frac{1}{n} \sumizn e^{\omega x_i})$ and $d_\omega(\bX)=\omega$. Then:
\begin{equation*}
\frac{\partial n_\omega(\bX)}{\partial \omega}=\frac{\sumizn x_i e^{\omega x_i}}{\sumizn e^{\omega x_i}}\quad \textrm{and} \quad \frac{\partial d_\omega(\bX)}{\partial \omega}=1\ ,
\label{e:beta}
\end{equation*}
and so:
\begin{equation*}
    \frac{\partial \mm_\omega(\bX)}{\partial \omega}=\frac{\frac{\partial n_\omega(\bX)}{\partial \omega}d_\omega(\bX) - n_\omega(\bX) \frac{\partial d_\omega(\bX)}{\partial \omega}}{d_\omega(\bX)^2} \ ,
\end{equation*}
ensuring differentiablity of the operator (Property~\ref{p:differentiable}).

\subsection{Averaging}

Because of the division by $\omega$ in the definition of $\mm_\omega$,
the parameter $\omega$ cannot be set to zero. However, we can examine the behavior of
$\mm_\omega$ as $\omega$ approaches zero and show that the operator computes an average in the limit.

Since both the numerator and
denominator go to zero as $\omega$ goes to zero, we will use
L'H\^{o}pital's rule and the derivative given in the previous section
%Equation~\ref{e:beta} 
to derive the value in the limit:
\begin{eqnarray*}
\lim_{\omega\rightarrow 0} \mm_\omega(\bX)
&=& \lim_{\omega\rightarrow 0} \frac{ \log (\frac{1}{n} \sumizn e^{\omega x_i})}{\omega} \\
&\stackrel{\text{L'H\^{o}pital}}{=}&  \lim_{\omega\rightarrow 0} \frac{\frac{1}{n} \sumizn x_i e^{\omega x_i}}{\frac{1}{n} \sumizn e^{\omega x_i}}\\
&=& \frac{1}{n} \sumizn x_i
= \mean(\bX)\ .
\end{eqnarray*}
That is, as $\omega$ gets closer to zero, $\mm_\omega(\bX)$ approaches the mean of the values in $\bX$.

\section{Maximum Entropy Mellowmax Policy}
\label{sec:max_ent_mellow_policy}
As described, $\mm_\omega$ computes a value for a list of numbers somewhere between its minimum and maximum. However, it is often useful to actually provide a probability distribution over the actions such that (1) a non-zero probability mass is assigned to each action, and (2) the resulting expected value equals the computed value. Such a probability distribution can then be used for action selection in algorithms such as SARSA.

In this section, we address the problem of identifying such a probability distribution as a maximum entropy problem---over all distributions that satisfy the properties above, pick the one that maximizes information entropy~\citep{cover2006,peters2010relative}. We formally define the maximum entropy mellowmax policy of a state $s$ as:
\begin{eqnarray}
&&\pi_{\rm mm}(s)=\argmin_\pi \sum_{a\in\mathcal{A}}\pi(a|s)\log\big(\pi(a|s)\big)\label{eq:convex_opt_problem} \\ 
&&\textrm{subject to}\ \Big\{  
\begin{array}{l}
        \sum_{a\in\mathcal{A}}\pi(a|s)\hat Q(s,a)=\mm_\omega(\hat Q(s,.)) \\
        \pi(a|s) \ge 0 \\
        \sum_{a\in\mathcal{A}}\pi(a|s)=1 \ . % & 
\end{array}\nonumber
\end{eqnarray}
Note that this optimization problem is convex and can be solved reliably using any numerical convex optimization library.

One way of finding the solution, which leads to an interesting policy form, is to use the method 
of Lagrange multipliers. Here, the Lagrangian is:
\begin{eqnarray*}
&&L(\pi,\lambda_1,\lambda_2)=\sum_{a\in\mathcal{A}}\pi(a|s)\log\big(\pi(a|s)\big)\\ &&-\lambda_1\big(\sum_{a\in\mathcal{A}}\pi(a|s)-1\big)\\
&&- \lambda_2\Big(\sum_{a\in\mathcal{A}}\pi(a|s)\hat Q(s,a)-\mm_\omega\big(\hat{Q}(s,.)\big)\Big)\ .\\
\end{eqnarray*}
Taking the partial derivative of the Lagrangian with respect to each $\pi(a|s)$ and setting them to zero, we obtain:
$$\frac{\partial L}{\partial \pi(a|s)}=\log\big(\pi(a|s)\big)+1-\lambda_1-\lambda_2\hat Q(s,a)=0\quad \forall\ a \in 
\mathcal{A}\ .$$
 
These $|\mathcal{A}|$ equations, together with the two linear constraints in (\ref{eq:convex_opt_problem}), form $|\mathcal{A}|+2$ equations to constrain the $|\mathcal{A}|+2$ variables $\pi(a|s)\ \forall a \in \mathcal{A}$ and the two Lagrangian multipliers $\lambda_1$ and $\lambda_2$. 
% As a result, we have a system of equations with $|\mathcal{A}|+2$ equations and $|\mathcal{A}|+2$ variables.

Solving this system of equations, the probability of taking an action under the maximum entropy mellowmax policy has the form:
\begin{equation*}
\pi_{mm}(a|s)=\frac{e^{\beta \hat Q(s,a)}}{\sum_{a\in\mathcal{A}}e^{\beta\hat Q(s,a)}}\quad \forall a \in \mathcal{A} \ ,
\label{policy:Boltzmann}
\end{equation*}
where $\beta$ is a value for which:
$$\sum_{a \in \mathcal{A}} e^{\beta \big(\hat Q(s,a)-\mm_\omega\hat Q(s,.)\big)}\big(\hat Q(s,a)-\mm_\omega\hat Q(s,.)\big)=0\ .$$
The argument for the existence of a unique root is simple. As $\beta\rightarrow \infty$ the term corresponding to the best action dominates, and so, the function is positive. Conversely, as $\beta\rightarrow-\infty$ the term corresponding to the action with lowest utility dominates, and so the function is negative. Finally, by taking the derivative, it is clear that the function is monotonically increasing, allowing us to conclude that there exists only a single root. Therefore, we can find $\beta$ easily using any root-finding algorithm. In particular, we use Brent's method \citep{brent2013algorithms} available in the Numpy library of Python.

This policy has the same form as Boltzmann softmax, but with a parameter $\beta$ whose value depends indirectly on $\omega$. 
This mathematical form arose not from the structure of $\mm_\omega$, but from maximizing the entropy. One way to view the use of the mellowmax operator, then, is as a form of Boltzmann policy with a temperature parameter chosen adaptively in each state to ensure that the non-expansion property holds. 

Finally, note that the SARSA update under the maximum entropy mellowmax policy could be thought of as a stochastic implementation of the GVI update under the $\mm_\omega$ operator:
\begin{eqnarray*}
\lefteqn{\E_{\pi_{mm}}\big[r+\gamma \hat{Q}(s',a')\big|s,a\big]=}\\ 
&&\sum_{s' \in \mathcal S}\mathcal{R}(s,a,s')+\gamma\mathcal{P}(s,a,s')\underbrace{\sum_{a'\in\mathcal A}\pi_{mm}(a'|s')\hat Q(s',a')\big]}_{\mm_{\omega}\big(\hat Q(s',.)\big)} 
\end{eqnarray*}
due to the first constraint of the convex optimization problem (\ref{eq:convex_opt_problem}). Because mellowmax is a non-expansion, SARSA with the maximum entropy mellowmax policy is guaranteed to converge to a unique fixed point. Note also that, similar to other variants of SARSA, the algorithm simply bootstraps using the value of the next state while implementing the new policy.
\section{Experiments on MDPs}
We observed that in practice computing mellowmax can yield overflow if the exponentiated values are large. In this case, we can safely shift the values by a constant before exponentiating them due to the following equality:
\begin{equation*}
	\frac{ \log (\frac{1}{n} \sumizn e^{\omega
    x_i})}{\omega}=c+\frac{ \log (\frac{1}{n} \sumizn e^{\omega
    (x_i-c)})}{\omega}\ .
\end{equation*}
A value of $c=\max_i x_i$ usually avoids overflow.

\begin{figure}
\centering
\includegraphics[width=0.8\columnwidth]{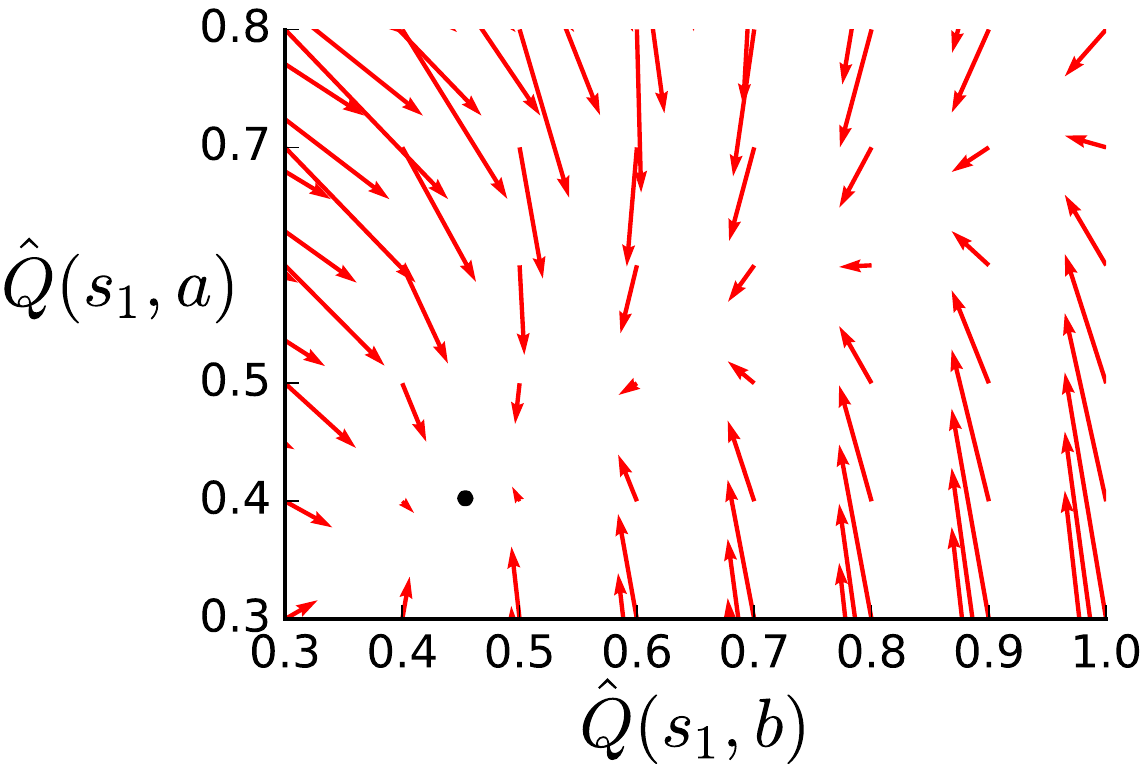}
\caption{GVI updates under $\mm_{\omega=16.55}$. The fixed point is unique, and all updates move quickly toward the fixed point.}
\label{f:vectorFieldBoltz1}
\end{figure}

\begin{figure}
\centering
\includegraphics[width=1\columnwidth]{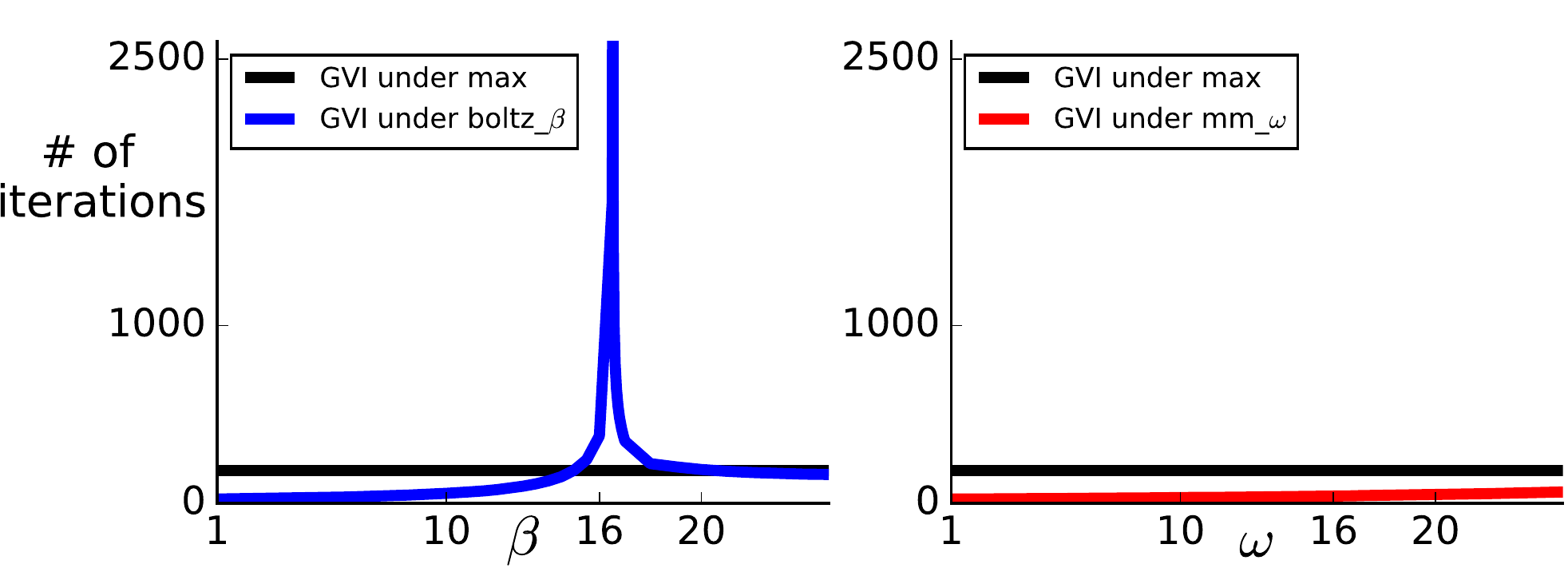}
\caption{Number of iterations before termination of GVI on the example MDP. GVI under $\mm_\omega$ outperforms the alternatives.}
\label{countsBoth}
\end{figure}

We repeat the experiment from Figure~\ref{f:vectorFieldBoltz} for mellowmax with $\omega=16.55$ to get a vector field. The result, presented in Figure \ref{f:vectorFieldBoltz1}, show a rapid and steady convergence towards the unique fixed point. As a result, GVI under $\mm_\omega$ can terminate significantly faster than GVI under $\boltz_\beta$, as illustrated in Figure~\ref{countsBoth}.

We present three additional experiments. The first experiment investigates the behavior of GVI with the softmax operators on randomly generated MDPs. The second experiment evaluates the softmax policies when used in SARSA with a tabular representation. The last experiment is a policy gradient experiment where a deep neural network, with a softmax output layer, is used to directly represent the policy.

\subsection{Random MDPs}

The example in Figure~\ref{f:ambig} was created carefully by hand. It is interesting to know whether such examples are likely to be encountered naturally. 
% We ran GVI on a collection of randomly generated MDPs.
To this end, we constructed 200 MDPs as follows:  We sampled $|\mathcal S|$ from $\{2, 3, ... ,10\}$ and $|\mathcal A|$ from $\{2, 3, 4, 5\}$ uniformly at random. We initialized the transition probabilities by sampling uniformly from $[0,.01]$. We then added to each entry, with probability 0.5, Gaussian noise with mean 1 and variance 0.1. We next added, with probability 0.1, Gaussian noise with mean 100 and variance 1. Finally, we normalized the raw values to ensure that we get a transition matrix. We did a similar process for rewards, with the difference that we divided each entry by the maximum entry and multiplied by 0.5 to ensure that $R_{\max}=0.5\ $.  

We measured the failure rate of GVI under $\boltz_\beta$ and $\mm_\omega$ by stopping GVI when it did not terminate in 1000 iterations.  We also computed the average number of iterations needed before termination. A summary of results is presented in the table below. Mellowmax outperforms Boltzmann based on the three measures provided below.
\begin{center}
\begin{tabular}{| p{1cm}|p{1.75cm}|p{1.75cm}| p{1.5cm}|}
\hline
   & MDPs, no terminate & MDPs, $>1$ fixed points & average iterations \\ \hline
  $\boltz_\beta$ & 8 of 200 & 3 of 200 & 231.65 \\ \hline
  $\mm_{\omega}$ & \textbf 0 & \textbf 0 & \textbf{201.32}\\ \hline
\end{tabular}
\end{center}
\subsection{Multi-passenger Taxi Domain}
\begin{figure}
\centering
  \includegraphics[clip,width=0.35\columnwidth]{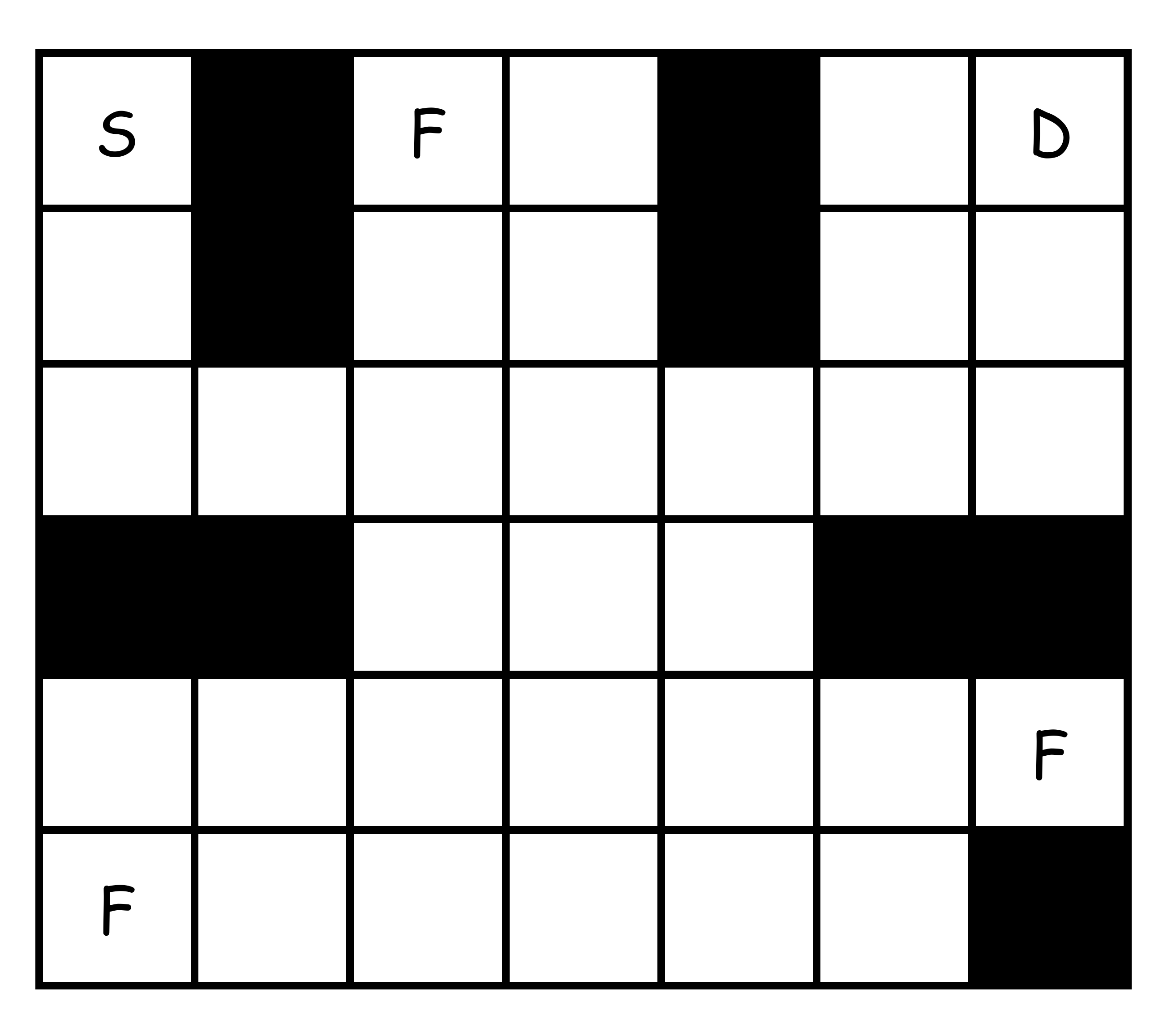}%
\caption{Multi-passenger taxi domain. The discount rate $\gamma$ is 0.99. Reward is $+1$ for delivering one passenger, $+3$ for two passengers, and $+15$ for three passengers. Reward is zero for all the other transitions. Here $F$, $S$, and $D$ denote passengers, start state, and destination respectively.}
\label{fig:Taxi}
\end{figure}

We evaluated SARSA on 
the multi-passenger taxi domain introduced by \citet{dearden98}. (See Figure~\ref{fig:Taxi}.)

One challenging aspect of this domain is that it admits many locally optimal policies.  Exploration needs to be set carefully to avoid either over-exploring or under-exploring the state space. Note also that Boltzmann softmax performs remarkably well on this domain, outperforming sophisticated Bayesian reinforcement-learning algorithms~\cite{dearden98}.
\begin{figure}
  \centering
  \includegraphics[clip,width=.5\columnwidth]{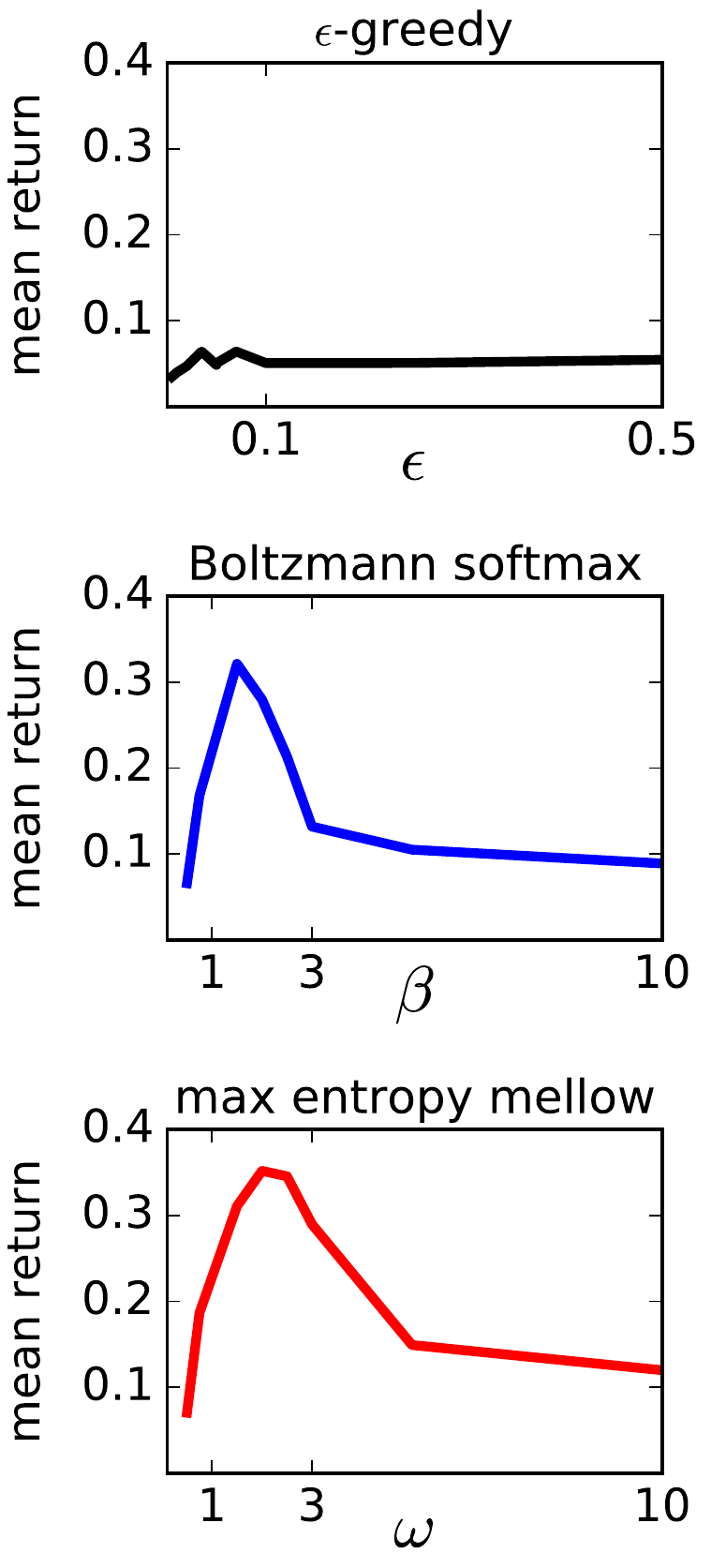}
\caption{Comparison on the multi-passenger taxi domain. Results are shown for different values of $\epsilon$, $\beta$, and $\omega$. For each setting, the learning rate is optimized. Results are averaged over 25 independent runs, each consisting of 300000 time steps.}
\label{fig:TaxiResults}
\end{figure}
As shown in Figure~\ref{fig:TaxiResults}, SARSA with the epsilon-greedy policy performs poorly. In fact, in our experiment, the algorithm rarely was able to deliver all the passengers. However, SARSA with Boltzmann softmax and SARSA with the maximum entropy mellowmax policy achieved significantly higher average reward. Maximum entropy mellowmax policy is no worse than Boltzmann softmax, here, suggesting that the greater stability does not come at the expense of less effective exploration.
\subsection{Lunar Lander Domain}

In this section, we evaluate the use of the maximum entropy mellowmax policy in the context of a policy-gradient algorithm. Specifically, we represent a policy by a neural network (discussed below) that maps from states to probabilities over actions. A common choice for the activation function of the last layer is the Boltzmann softmax policy. In contrast, we can use maximum entropy mellowmax policy, presented in Section~\ref{sec:max_ent_mellow_policy}, by treating the inputs of the activation function as $\hat Q$ values.

\begin{figure}
\centering
  \includegraphics[clip,width=.35\columnwidth]{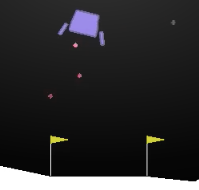}%
\caption{A screenshot of the lunar lander domain.}
\label{fig:lunar_lander}
\end{figure}

We used the lunar lander domain, from OpenAI Gym \cite{open_AI} as our benchmark. A screenshot of the domain is presented in Figure \ref{fig:lunar_lander}. This domain has a continuous state space with 8 dimensions, namely x-y coordinates, x-y velocities, angle and angular velocities, and leg-touchdown sensors. There are 4 discrete actions to control 3 engines. The reward is +100 for a safe landing in the designated area, and $-100$ for a crash. There is a small shaping reward for approaching the landing area. Using the engines results in a negative reward. An episode finishes when the spacecraft crashes or lands. Solving the domain is defined as maintaining mean episode return higher than 200 in 100 consecutive episodes.

\begin{figure}
\centering
   \includegraphics[width=.75\columnwidth]{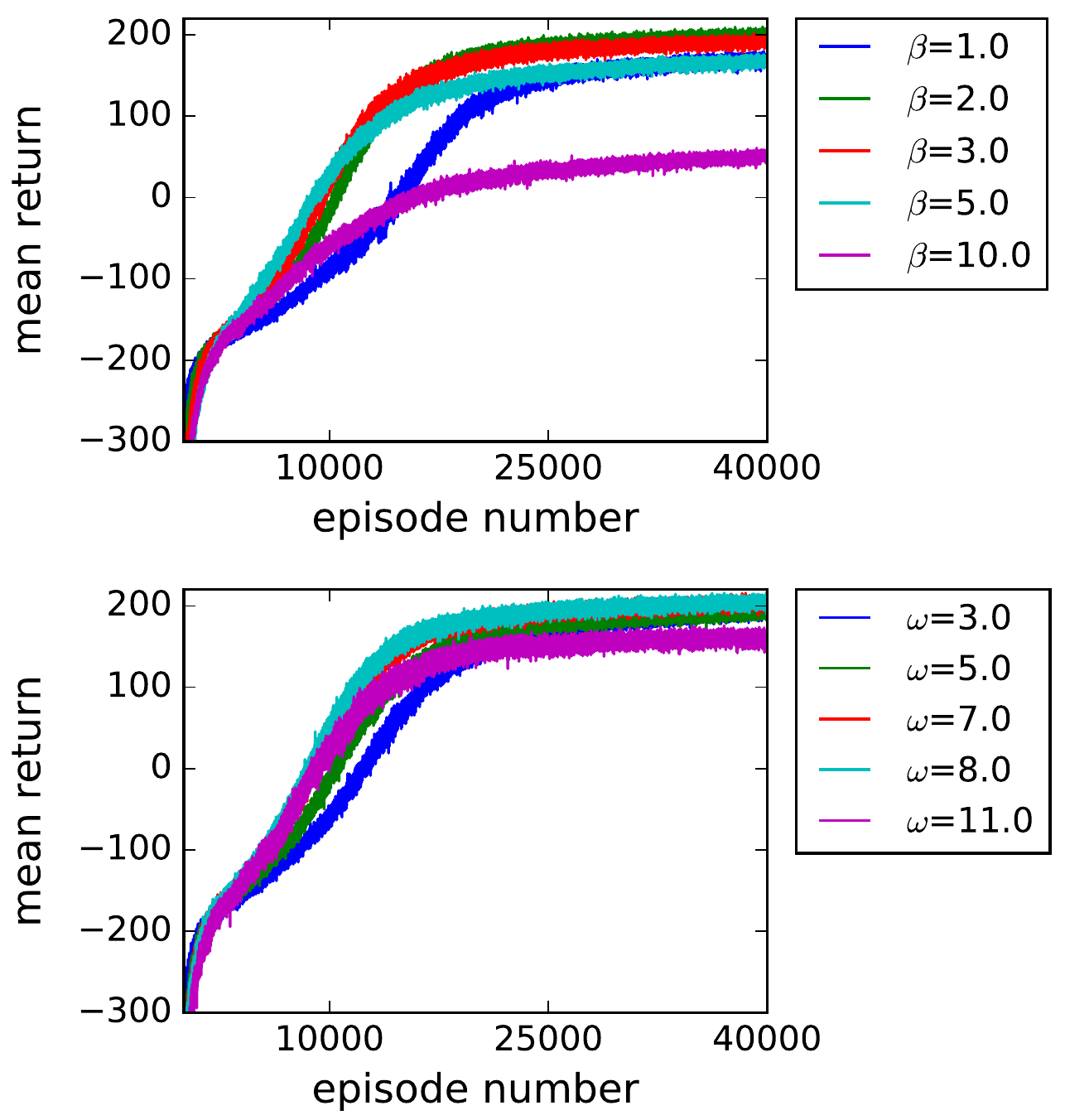}
   \includegraphics[width=0.9\columnwidth]{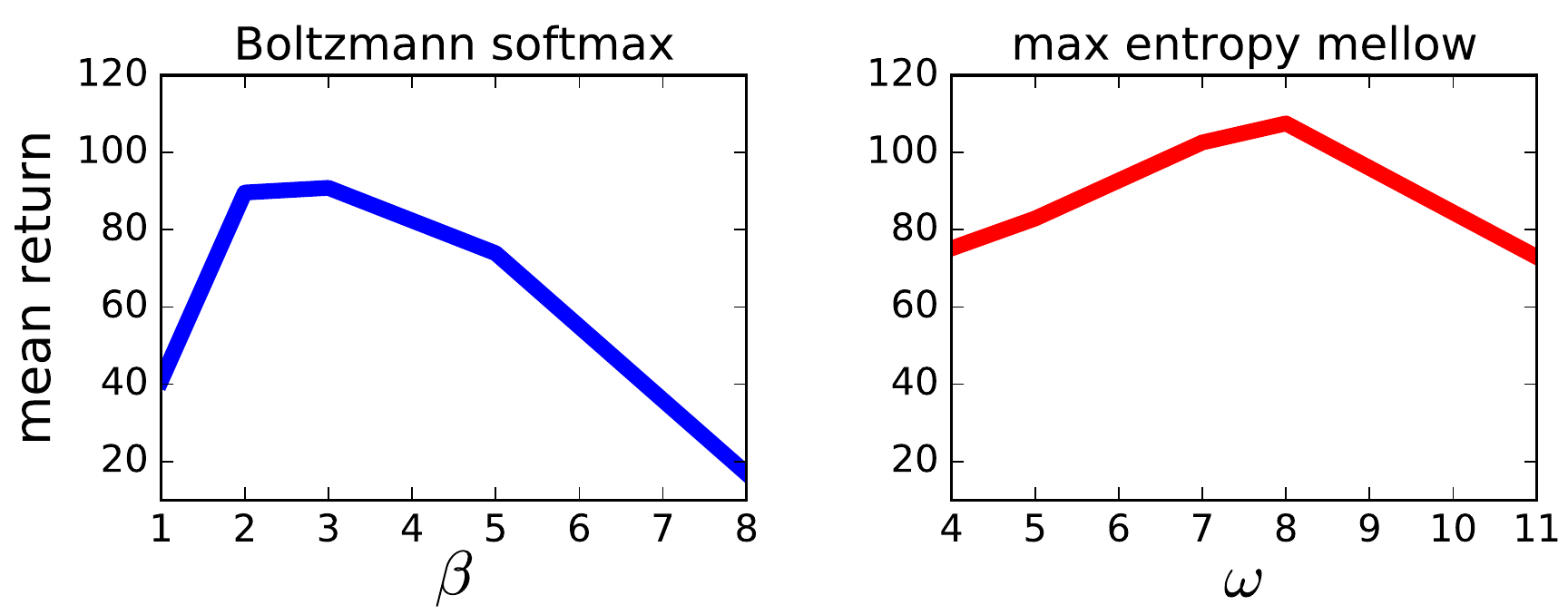}
\caption{Comparison of Boltzmann (top) and maximum entropy mellowmax (middle) in Lunar Lander. Mean return over all episodes (bottom). Results are 400-run averages.}
\label{fig:lunar_lander_learning_curves}
\end{figure}

The policy in our experiment is represented by a neural network with a hidden layer comprised of 16 units with RELU activation functions, followed by a second layer with 16 units and softmax activation functions. We used REINFORCE to train the network. A batch episode size of 10 was used, as we had stability issues with smaller episode batch sizes. We used the Adam algorithm \cite{kingma2014adam} with $\alpha=0.005$ and the other parameters as suggested by the paper. We used Keras \cite{chollet2015keras} and Theano \cite{team2016theano} to implement the neural network architecture. 
For each softmax policy, we present in Figure \ref{fig:lunar_lander_learning_curves} the learning curves for different values of their free parameter. We further plot average return over all 40000 episodes. Mellowmax outperforms Boltzmann at its~peak.
\section{Related Work}
Softmax operators play an important role in sequential decision-making algorithms.

In model-free reinforcement learning, they can help strike a balance between exploration (mean) and exploitation (max). Decision rules based on epsilon-greedy and Boltzmann softmax, while very simple,
often perform surprisingly well in practice, even outperforming more
advanced exploration techniques~\citep{kuleshov14} that require significant approximation for complex domains.
When learning ``on policy'', exploration steps can~\citep{rummery94b} and perhaps should~\citep{john94} become part of the value-estimation process itself. On-policy algorithms like SARSA can be made to converge to optimal behavior in the limit when the exploration rate and the update operator is gradually moved toward $\max$~\citep{singh00}. Our use of softmax in learning updates reflects this point of view and shows that the value-sensitive behavior of Boltzmann exploration can be maintained even as updates are made stable. 

Analyses of the behavior of human subjects in choice experiments very
frequently use softmax. Sometimes referred to in the
literature as logit choice~\citep{stahl94}, it forms an important part of the most accurate predictor of human decisions in normal-form games~\citep{wright10},
quantal level-$k$ reasoning (QLk). Softmax-based fixed points play a
crucial role in this work. As such, mellowmax could potentially make a good replacement.

Algorithms for inverse reinforcement learning (IRL), the problem of inferring reward functions from observed behavior~\citep{ng00}, frequently use a Boltzmann operator to avoid assigning zero probability to non-optimal actions and hence assessing an observed sequence as impossible. Such methods include Bayesian IRL~\citep{ramachandran07}, natural gradient IRL~\citep{neu07}, and maximum likelihood IRL~\citep{babes11}. Given the recursive nature of value defined in these problems, mellowmax could be a more stable and efficient choice.

In linearly solvable MDPs~\cite{todorov2006linearly}, an operator similar to mellowmax
emerges when using an alternative characterization for cost of action
selection in MDPs. Inspired by this
work~\citet{fox2015taming} introduced an off-policy G-learning
algorithm that uses the operator to perform value-function
updates. Instead of performing off-policy updates, we introduced a convergent variant of SARSA with Boltzmann policy and a state-dependent temperature parameter. This is in contrast to \citet{fox2015taming} where an epsilon greedy behavior policy is used.

\section{Conclusion and Future Work}

We proposed the mellowmax operator as an alternative to the Boltzmann softmax operator. We showed that mellowmax has several desirable properties and that it works favorably in practice. Arguably, mellowmax could be used in place of Boltzmann throughout reinforcement-learning research.

A future direction is to analyze the fixed point of planning, reinforcement-learning, and game-playing algorithms when using the mellowmax operators. In particular, an interesting analysis could be one that bounds the sub-optimality of the fixed points found by GVI.

An important future work is to expand the scope of our theoretical understanding to the more general function approximation setting, in which the state space or the action space is large and abstraction techniques are used. Note that the importance of non-expansion in the function approximation case is well-established. \cite{gordon1995stable}

Finally, due to the convexity of mellowmax~\citep{boyd2004convex}, it
is compelling to use it in a gradient-based algorithm in the context of sequential decision making. IRL is a natural candidate given the popularity of softmax in this setting.
\section{Acknowledgments}
The authors gratefully acknowledge the assistance of George D. Konidaris, as well as anonymous ICML reviewers for their outstanding feedback.
\bibliography{mlittman,add}
\bibliographystyle{icml2017}

\end{document}